\documentclass[11pt]{article}

\usepackage[preprint]{acl}

\usepackage{times}
\usepackage{latexsym}
\usepackage{amsmath}
\usepackage{amssymb}
\usepackage{booktabs}
\usepackage{graphicx}
\usepackage{multirow}
\usepackage{colortbl}
\usepackage{adjustbox}
\usepackage{algorithm}
\usepackage{algpseudocode}
\usepackage{enumitem}
\usepackage{float}
\usepackage{capt-of}
\usepackage{caption}
\usepackage{pgfplots}
\pgfplotsset{compat=1.18}

\pgfplotsset{compat=1.18}

\usepackage{enumitem}
\setlist[itemize]{itemsep=0pt, parsep=0pt, topsep=1pt, partopsep=0pt}
\setlist[enumerate]{itemsep=0pt, parsep=0pt, topsep=1pt, partopsep=0pt}

\usepackage[T1]{fontenc}

\usepackage[utf8]{inputenc}

\usepackage{microtype}

\usepackage{inconsolata}

\usepackage{graphicx}

\title{DynaCF: Mitigating Shortcut Learning in Reward Models via Dynamic Counterfactual Sensitivity}

\author{
Fengyuan Liu$^{1}$\thanks{Equal contribution.} \,
Yongliang Miao$^{1}$\footnotemark[1] \,
Zirui He$^{2}$ \,
Yanguang Liu$^{2}$ \,
\textbf{Fei Sun}$^{3}$ \,
\textbf{Mengnan Du}$^{1}$\thanks{Corresponding author.} \\
$^{1}$The Chinese University of Hong Kong, Shenzhen \quad \\
$^{2}$New Jersey Institute of Technology \quad 
$^{3}$Institute of Computing Technology, CAS \\
\texttt{\{liuferry708, r130026108\}@gmail.com} \quad
\texttt{\{zh296, yanguang.liu\}@njit.edu} \quad \\
\texttt{mengnandu@cuhk.edu.cn}
}

\begin{document}
\maketitle
\begin{abstract}

Reward models trained from pairwise preferences often exploit superficial shortcut cues rather than learning true response quality. We propose DynaCF, a dynamic reweighting framework for mitigating shortcut learning in reward model training. Unlike static shortcut heuristics, DynaCF measures shortcut sensitivity online during optimization by applying semantics-preserving counterfactual perturbations and tracking the resulting margin shifts and preference flips under the current model. Samples with higher shortcut sensitivity are dynamically downweighted in the Bradley–Terry objective, encouraging the model to rely less on superficial patterns and more on task-relevant preference signals. Extensive experiments show that DynaCF consistently improves robustness in preference modeling.
\end{abstract}

\section{Introduction}

\begin{figure*}[t!]
    \centering
    \includegraphics[width=\textwidth]{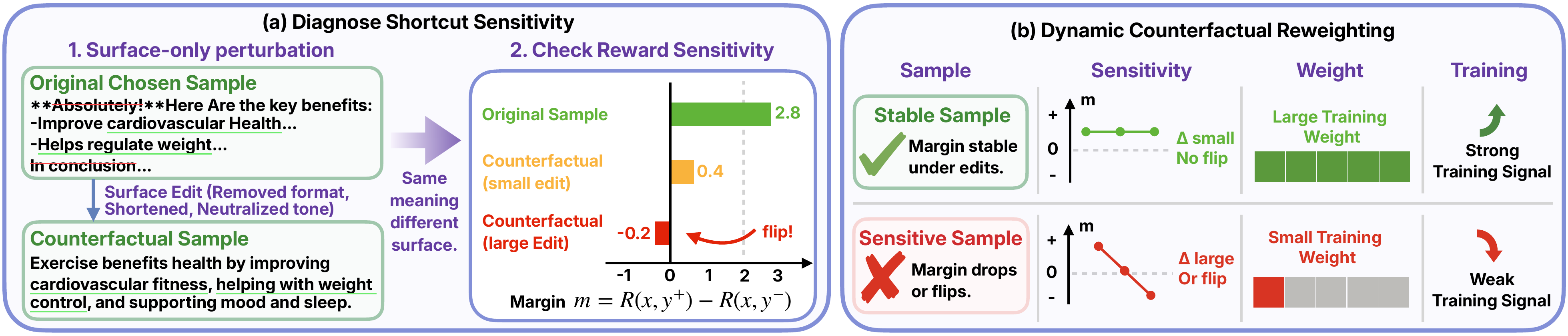}
    \caption{
Overview of DynaCF. DynaCF constructs semantics-preserving counterfactual variants of the chosen response, evaluates shortcut sensitivity with the current reward model, and converts margin shifts and preference flips into dynamic weights for the original Bradley--Terry objective.
}
    \label{fig:framework}
\end{figure*}

Reward models play a central role in aligning language models with human preferences and are commonly trained from pairwise preference data~\cite{christiano2017deep,ziegler2019fine,stiennon2020learning,ouyang2022training,bai2022training,rafailov2023direct}. However, preference pairs often contain not only task-relevant quality differences but also superficial cues such as length, formatting, verbosity, or template-like expressions. As a result, reward models may exploit these shortcuts instead of learning intended preference signals~\cite{geirhos2020shortcut,singhal2023long,shen2023loose,park2024disentangling,dubois2024length}, undermining robustness by encouraging decisions based on superficial patterns.

A central difficulty is that shortcut risk is not simply a static property of a dataset or a predefined surface feature. For example, a longer or better formatted chosen response is not necessarily harmful if the model's preference remains stable when these surface properties are neutralized. Conversely, a seemingly benign preference pair may become shortcut-sensitive if the current reward model relies heavily on superficial cues to maintain its margin. Therefore, whether a sample induces shortcut learning depends on the model's current behavior and may change across training stages. This observation suggests that shortcut mitigation should move beyond static filtering or augmentation and instead diagnose, during training, how sensitive the current reward model is to semantics-preserving surface perturbations.

Motivated by this view, we propose \textbf{DynaCF}, a framework that leverages Dynamic Counterfactual Sensitivity to diagnose shortcut-prone preference pairs and reweight reward-model training. As illustrated in Figure~\ref{fig:framework}, DynaCF constructs semantics-preserving counterfactual variants of the chosen response while keeping the prompt and rejected response fixed. These counterfactuals are used to probe the reward model: large reward-margin shifts or preference flips under surface-only changes indicate high counterfactual shortcut sensitivity. DynaCF then converts this sensitivity into a step-local sample weight for the original Bradley--Terry objective, downweighting shortcut-sensitive updates without deleting samples or treating counterfactual responses as additional preference labels.Experiments on RM-Bench, RewardBench, and RewardBench2~\cite{liu2024rmbench,lambert2024rewardbench,malik2025rewardbench2} show that DynaCF improves reward-model robustness over Bradley--Terry training, with RM-Bench Hard gains of $+8.7$ and $+4.6$ points and Safety gains of $+8.2$ and $+1.1$ on Qwen3-4B and Qwen3-8B, respectively.

\section{Methodology}

In this section, we briefly introduce reward model training and then present DynaCF, a dynamic counterfactual reweighting framework that assesses reliance on superficial features and downweights shortcut-sensitive samples at each step.

\subsection{Preliminary and Task Definition}

Given a preference dataset 
$\mathcal{D}=\{(x_i,y_i^+,y_i^-)\}_{i=1}^N$, where $x_i$
 is the prompt, $y_i^+$ is the chosen response, and $y_i^-$ is the rejected response, a reward model $R_\theta$ assigns a scalar reward to each prompt-response pair. Standard Bradley--Terry training~\cite{bradley1952rank} maximizes the probability that the chosen response receives a higher reward than the rejected one, which is optimized by minimizing
\begin{equation}
\label{eq:preliminary-bt-loss}
\mathcal{L}_{\mathrm{bt}}(\theta)
=
-\frac{1}{N}\sum_{i=1}^{N}\log\sigma(m_i),
\end{equation}
where the reward margin between the chosen and rejected responses is defined as
\begin{equation}
\label{eq:preliminary-margin}
m_i=R_\theta(x_i,y_i^+)-R_\theta(x_i,y_i^-).
\end{equation}
While effective, relying solely on pairwise preference data with the Bradley--Terry objective can encourage shortcut learning, as differences in superficial properties between responses may be spuriously aligned with preference labels.

\subsection{Counterfactual Shortcut Sensitivity}

Counterfactual Shortcut Sensitivity measures how sensitive a reward model is to shortcut cues in a preference pair. We construct semantics-preserving counterfactual variants of the chosen response while keeping the prompt and rejected response fixed. For sample $i$, the original pair is
$
(x_i, y_i^+, y_i^-)
$
and the counterfactual pair is
$
(x_i, y_{i,k}^{\mathrm{cf}}, y_i^-)
$,
where $y_{i,k}^{\mathrm{cf}}$ is the $k$-th valid counterfactual. These variants preserve task-relevant content while neutralizing surface properties such as formatting, verbosity, tone, or template-like expressions. Given a reward model $R_\theta$, we compute the original and counterfactual margins as follows:
\begin{equation}
\label{eq:counterfactual-margins}
\begin{aligned}
m_i(\theta)
&= R_\theta(x_i, y_i^+) - R_\theta(x_i, y_i^-), \\
m_{i,k}^{\mathrm{cf}}(\theta)
&= R_\theta(x_i, y_{i,k}^{\mathrm{cf}}) - R_\theta(x_i, y_i^-).
\end{aligned}
\end{equation}
We derive the margin change $\Delta_{i,k}(\theta)$ and preference-flip indicator $F_{i,k}(\theta)$:
\begin{equation}
\label{eq:counterfactual-diagnostics}
\begin{aligned}
\Delta_{i,k}(\theta)
&= \left|m_i(\theta)-m_{i,k}^{\mathrm{cf}}(\theta)\right|, \\
F_{i,k}(\theta)
&= \mathbb{I}\left[m_{i,k}^{\mathrm{cf}}(\theta)<0\right].
\end{aligned}
\end{equation}
The sensitivity score is
\begin{equation}
\label{eq:counterfactual-sensitivity}
s_{\mathrm{cf},i}(\theta)
=
\frac{1}{K_i}\sum_{k=1}^{K_i}\Delta_{i,k}(\theta)
\!+\!
\mu_{\mathrm{flip}}
\frac{1}{K_i}\sum_{k=1}^{K_i}F_{i,k}(\theta),
\end{equation}
where $K_i$ is the number of valid counterfactuals and $\mu_{\mathrm{flip}}$ controls the flip penalty. Larger $s_{\mathrm{cf},i}(\theta)$ indicates less stable reward-model preferences under surface-only changes.

If no valid counterfactual is available, we set
\begin{equation}
\label{eq:counterfactual-no-valid}
K_i=0 \Rightarrow s_{\mathrm{cf},i}(\theta)=0
\end{equation}

\begin{table*}[!t]
  \centering
  \footnotesize
  \setlength{\tabcolsep}{3.0pt}
  \renewcommand{\arraystretch}{1.0}
  \begin{adjustbox}{max width=\textwidth}
  \begin{tabular}{@{}l*{12}{c}@{}}
  \toprule
  \multirow{2}{*}{\raisebox{-0.35ex}{\textbf{Model}}}
  & \multicolumn{8}{c}{\textbf{RM-Bench}}
  & \multicolumn{2}{c}{\textbf{RewardBench}}
  & \multicolumn{2}{c}{\textbf{RewardBench2}} \\
  \cmidrule(lr){2-9}
  \cmidrule(lr){10-11}
  \cmidrule(lr){12-13}
  & Chat & Math & Code & Safety
  & Easy & Normal & Hard & Overall
  & Safety & Overall
  & Safety & Overall \\
  \midrule

  \rowcolor{gray!15}
  \multicolumn{13}{l}{\textbf{External Baselines}} \\
  Skywork-Reward-Gemma-2-27B
  & 69.5 & 54.7 & 53.2 & 91.9 & 78.0 & 69.2 & 54.9 & 67.3
  & 91.9 & 93.8 & 94.2 & 75.8 \\
  Skywork-Reward-Llama-3.1-8B
  & 69.5 & 60.6 & 54.5 & 95.7 & 89.0 & 74.7 & 46.6 & 70.1
  & 90.8 & 92.5 & 93.3 & 73.1 \\
  URM-Llama-3.1-8B
  & 71.2 & 61.8 & 54.1 & 93.1 & 84.0 & 73.2 & 53.0 & 70.0
  & 91.1 & 92.9 & 91.8 & 73.9 \\
  \midrule
  
  \rowcolor{gray!15}
  \multicolumn{13}{l}{\textbf{Qwen3-4B}} \\
  w/ Bradley--Terry
  & 65.5 & 67.3 & 62.6 & 80.5 & 91.3 & 74.4 & 41.3 & 69.0
  & 78.7 & 83.2 & 73.3 & 63.5 \\
  \rowcolor{blue!8}
  w/ DynaCF
  & \textbf{71.2} & \textbf{69.2} & \textbf{63.4} & \textbf{88.7}
  & \textbf{91.1} & \textbf{78.2} & \textbf{50.0} & \textbf{73.1}
  & \textbf{82.7} & \textbf{85.5} & \textbf{79.4} & \textbf{66.1} \\
  \midrule
  
  \rowcolor{gray!15}
  \multicolumn{13}{l}{\textbf{Qwen3-8B}} \\
  w/ Bradley--Terry
  & 68.1 & 69.8 & 60.8 & 90.2 & 90.3 & 76.6 & 49.8 & 72.2
  & 82.8 & 86.9 & 78.9 & 67.9 \\
  \rowcolor{blue!8}
  w/ DynaCF
  & \textbf{72.1} & \textbf{71.0} & \textbf{62.6} & \textbf{91.3}
  & \textbf{90.4} & \textbf{78.0} & \textbf{54.4} & \textbf{74.2}
  & \textbf{84.3} & \textbf{87.5} & \textbf{81.8} & \textbf{68.0} \\
  \bottomrule
  \end{tabular}
  \end{adjustbox}
  \caption{Evaluation results on RM-Bench, RewardBench, and RewardBench2. RM-Bench is reported with domain-level scores, difficulty-level scores, and overall accuracy; RewardBench and RewardBench2 are reported with safety and overall scores following official evaluation settings.}
  \label{tab:main_results}
\end{table*}

\subsection{Dynamic Counterfactual Reweighting}

The sensitivity score provides a model-dependent signal for shortcut risk. Rather than assigning fixed sample weights from static surface features, DynaCF recomputes sensitivity with the current reward model and reweights the original Bradley--Terry loss online. At step $t$, each sample in batch $B_t$ is evaluated by $R_{\theta_t}$, giving $s_{\mathrm{cf},i}(t)=s_{\mathrm{cf},i}(\theta_t)$ and the step-local weight:
\begin{equation}
\label{eq:dynamic-weight}
w_i(t)
=
\max\left(w_{\min},\,1-\gamma s_{\mathrm{cf},i}(t)\right),
\end{equation}
where $\gamma$ controls the strength of downweighting and $w_{\min}$ prevents samples from being completely removed.
The final training objective is
\begin{equation}
\label{eq:dynamic-objective}
\mathcal{L}_{\mathrm{DynaCF}}(t)
=
\frac{1}{|B_t|}
\sum_{i\in B_t}
w_i(t)\left[-\log\sigma(m_i(t))\right]
\end{equation}
Because weights depend on the current model, the same sample can be weighted differently across training stages. Counterfactuals are used only as diagnostic probes: DynaCF does not delete samples or treat counterfactual responses as additional preference labels. The full training procedure is provided in Appendix~\ref{sec:appendix_dynacf_algorithm}.

\subsection{Counterfactual Construction}

We construct counterfactuals for each sample with domain-aware rules rather than free-form LLM edits. For normal, math, and code examples, simple substitutions and rule-based normalization modify surface cues while preserving the chosen response's main content. Lightweight validity checks keep only sufficiently close counterfactuals; math/code examples must also preserve key structured elements such as numbers. Samples without valid counterfactuals retain their original weight. Details are in Appendix~\ref{sec:appendix_counterfactual_design}.

\section{Experiments}

In this section, we evaluate the extent to which DynaCF mitigates shortcut learning during reward-model training and improves preference modeling over standard Bradley--Terry training across multiple benchmarks and model scales.

\subsection{Experimental Settings}

\paragraph{Models and Data.}
We instantiate reward models with Qwen3-4B and Qwen3-8B~\cite{yang2025qwen3} to examine whether DynaCF remains effective across different backbone capacities. Both models are trained on the HelpSteer3 preference split~\cite{wang2025helpsteer3}. We evaluate reward-model generalization with three complementary benchmarks: RM-Bench, RewardBench, and RewardBench2~\cite{liu2024rmbench,lambert2024rewardbench,malik2025rewardbench2}. 

\paragraph{Implementation Details.}
We train all models with learning rate $4.0{\times}10^{-5}$ and LR warmup ratio $0.03$. The maximum sequence length is set to $2048$ for Qwen3-4B and $4096$ for Qwen3-8B. For DynaCF, All counterfactual samples are constructed offline before training. We set a reweighting warmup ratio of $0.05$, with $\gamma=1.0$ and $w_{\min}=0.1$. All ablation experiments are conducted with Qwen3-4B. More training details are provided in Appendix~\ref{sec:appendix_training_details}. 

\paragraph{Baselines.}

We compare DynaCF with: (i) vanilla baselines, which are reward models trained with the Bradley--Terry objective; (ii) external baselines, which are open-source reward models collected from public leaderboards and recent reward-modeling work~\cite{liu2024skywork,liu2025skyworkv2,yang2024grm,lou2024urm,wang2024interpretable}, used to contextualize the performance of DynaCF.

\subsection{Main Results}

\noindent \textbf{Difficulty Amplifies Shortcut Reliance.}
A granular examination of Table~\ref{tab:main_results} shows that DynaCF consistently improves performance on hard reasoning and safety-sensitive settings, where superficial cues are less reliable indicators of true response quality. On the RM-Bench Hard split, the Bradley--Terry baseline drops to $41.3\%$ for Qwen3-4B, suggesting an over-reliance on shortcut patterns that fail against hard-negative distractors. By dynamically downweighting shortcut-sensitive samples through online counterfactual sensitivity estimation, DynaCF improves the Hard split by $+8.7$ points for Qwen3-4B ($41.3\%\rightarrow50.0\%$) and $+4.6$ points for Qwen3-8B ($49.8\%\rightarrow54.4\%$), while largely preserving performance on Easy examples. 

\noindent \textbf{Robustness Gains in Safety-Critical Preferences.}
DynaCF also yields substantial gains on safety-related evaluations. On Qwen3-4B, DynaCF improves Safety performance by $+8.2$ on RM-Bench, $+4.0$ on RewardBench, and $+6.1$ on RewardBench2, indicating improved robustness against superficial stylistic or formatting cues in safety-critical preference pairs. Compared with the larger Qwen3-8B, the smaller Qwen3-4B benefits more consistently across benchmarks, suggesting that smaller reward models are more vulnerable to shortcut learning and therefore gain more from dynamic counterfactual reweighting.


\begin{figure}[t]
  \centering
  \IfFileExists{figures/shortcut_margin_instability.pdf}{
    \includegraphics[width=\columnwidth]{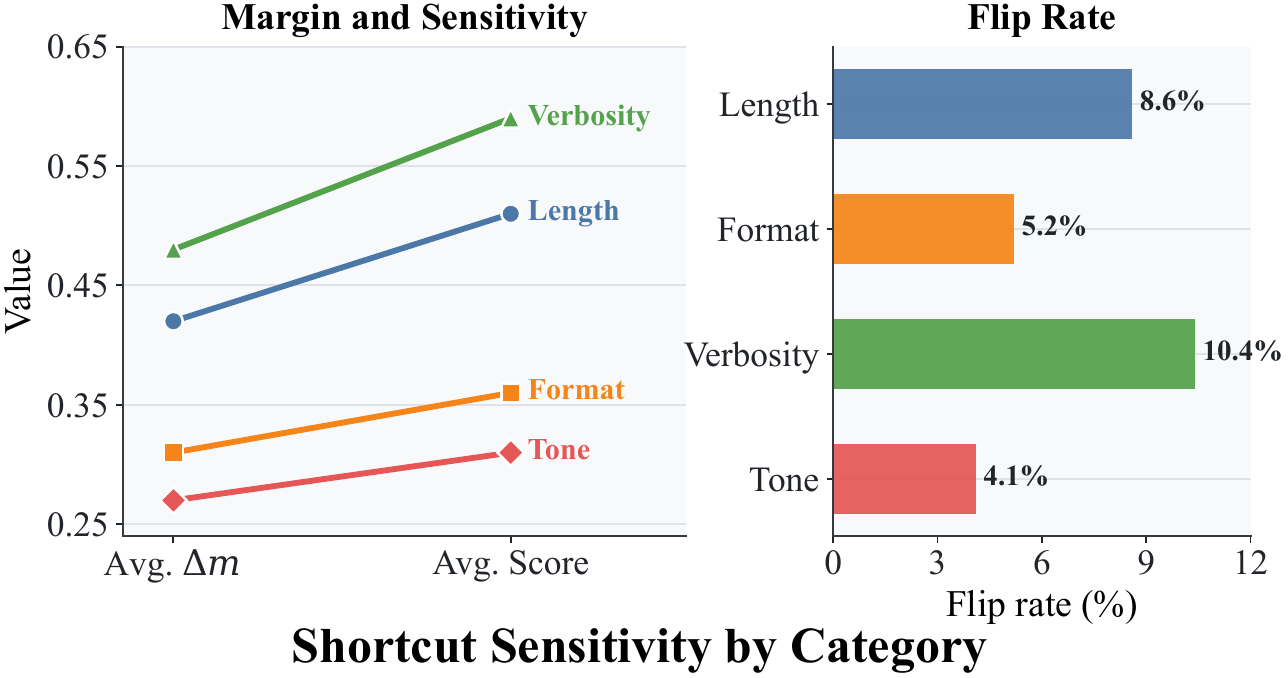}
  }{
    \fbox{\parbox[c][3.0cm][c]{0.95\columnwidth}{\centering figures/shortcut\_margin\_instability.pdf}}
  }
  \caption{Diagnostic statistics for reward-margin instability under different shortcut categories. Larger values indicate stronger sensitivity to the shortcut.}
  \label{fig:shortcut_margin_instability}
\end{figure}

\subsection{Analysis and Discussion}

\begin{figure}[t]
    \centering
    \includegraphics[width=\columnwidth]{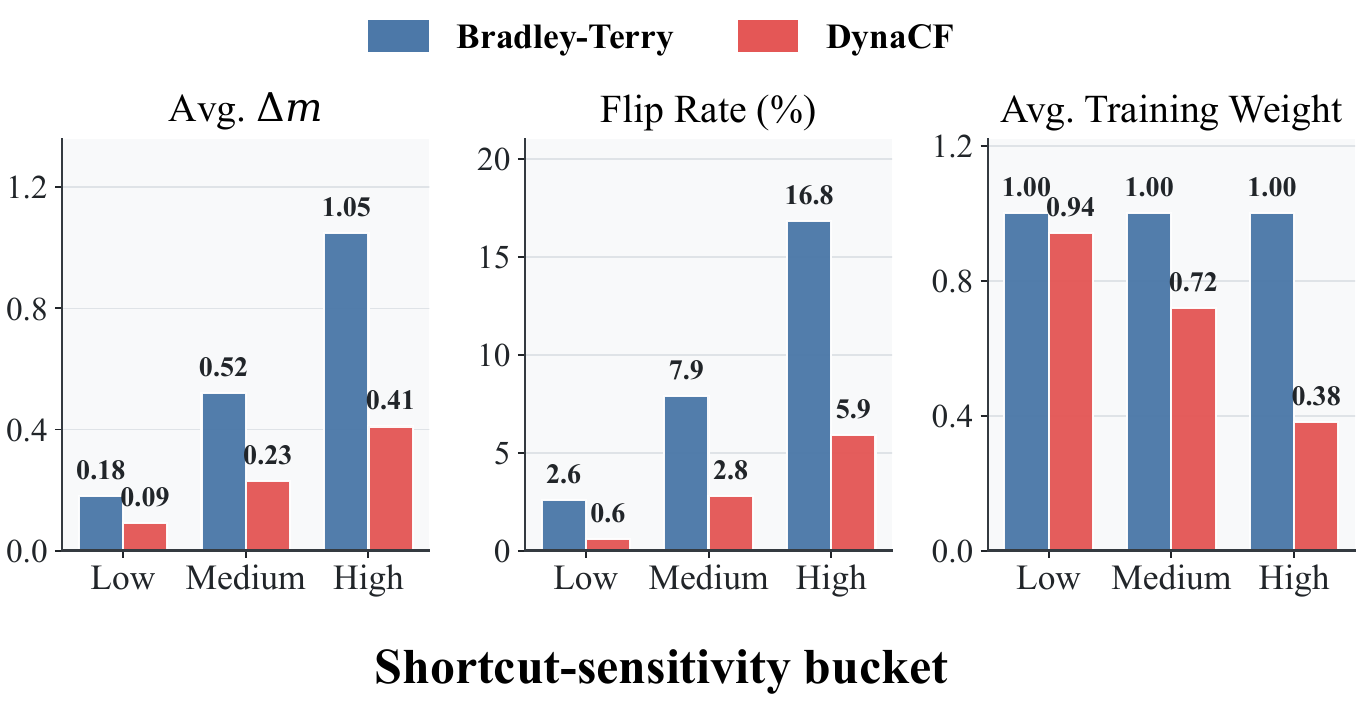}
    \caption{Shortcut sensitivity across low, medium, and high shortcut-sensitivity groups.
    Lower margin change and flip rate indicate better robustness to surface-only perturbations, while lower average weight indicates stronger downweighting by DynaCF.}
    \label{fig:shortcut_reweighting_effect}
\end{figure}

\noindent \textbf{Shortcut Analysis.}
We consider a shortcut cue to be detrimental when normalizing it preserves task-relevant content but substantially changes the reward margin. Figure~\ref{fig:shortcut_margin_instability} shows that several surface categories induce large margin shifts and non-trivial preference flips, suggesting that the Bradley--Terry baseline often anchors decisions on presentation cues rather than stable semantic quality. High-sensitivity samples are therefore the most likely to inject misleading gradients during reward-model training. Figure~\ref{fig:shortcut_reweighting_effect} shows that DynaCF reduces both margin changes and flips across sensitivity groups, with the strongest robustness gain on high-sensitivity samples. The learned weights also decrease with sensitivity, indicating that DynaCF preserves most low-sensitivity preference signal while selectively dampening updates from examples easiest to explain through superficial cues.

\noindent \textbf{Dynamic Analysis.} 
We vary how often shortcut sensitivity is recomputed. As shown in Table~\ref{tab:score_update_frequency}, static scores computed before training perform worse than the Bradley--Terry baseline, especially on the Hard split. This indicates that stale sensitivity estimates are not merely uninformative but can be actively harmful: before training, the reward model has not yet formed the decision boundary under which shortcut reliance emerges, so early counterfactual scores may misidentify reliable preference examples as shortcut-sensitive or fail to identify samples that become problematic later. Such incorrect weights distort the Bradley--Terry objective and suppress useful preference signals. In contrast, recomputing sensitivity after warmup provides a more reliable diagnostic signal, and more frequent updates further improve RM-Bench difficulty-level and overall scores by tracking emerging shortcut-sensitive samples during optimization. The full DynaCF setting used in the experiments adopts this warmup-based online recomputation strategy, with sensitivity scores updated throughout training; its result corresponds to the Qwen3-4B DynaCF entry in Table~\ref{tab:main_results}.


\begin{table}[t]
  \centering
  \footnotesize
  \setlength{\tabcolsep}{4pt}
  \renewcommand{\arraystretch}{1.0}
  \begin{adjustbox}{center,max width=1.08\columnwidth}
  \begin{tabular}{@{}lcccc@{}}
  \toprule
  \textbf{Score Update} & \textbf{Easy} & \textbf{Normal} & \textbf{Hard} & \textbf{Overall} \\
  \midrule
  BT baseline & 91.3 & 74.4 & 41.3 & 69.0 \\
  Pre-train & 88.7 & 72.1 & 38.5 & 66.4 \\
  Warmup once & 90.0 & 73.2 & 40.2 & 67.7 \\
  Freq. 0.20 & 90.6 & 75.2 & 44.1 & 70.6 \\
  Freq. 0.10 & 90.8 & 76.1 & 46.3 & 71.8 \\
  Freq. 0.05 & 91.0 & 77.2 & 48.4 & 72.6 \\
  Full DynaCF & \textbf{91.1} & \textbf{78.2} & \textbf{50.0} & \textbf{73.1} \\
  \bottomrule
  \end{tabular}
  \end{adjustbox}
  \caption{Dynamic analysis of score computation frequency using Qwen3-4B on RM-Bench. Frequency denotes recomputation after warmup.}
  \label{tab:score_update_frequency}
\end{table}


\section{Conclusions and Future Work}

We presented DynaCF, a dynamic counterfactual reweighting method that mitigates shortcut learning in reward model training. 
Experiments with Qwen3 reward models on HelpSteer3 demonstrate consistent gains over standard Bradley--Terry training across RM-Bench, RewardBench, and RewardBench2, especially on hard and safety-oriented evaluations. In future, we plan to improve counterfactual construction, extending the approach to broader reward-modeling scenarios.

\section*{Limitations}


Our work has several limitations. First, DynaCF relies on the quality of counterfactual construction. Although our rule-based transformations and lightweight validity checks are designed to preserve task-relevant content, the resulting counterfactuals may still occasionally alter useful information or fail to fully remove the targeted shortcut cues. Improving counterfactual generation and validation is therefore an important direction for future work. Second, our experiments are limited to text-based reward models and text preference benchmarks. We have not yet evaluated DynaCF on multimodal reward models, process reward models, or reward models used in interactive training and deployment settings, leaving its generality in these scenarios to future investigation. Third, our counterfactual probing is applied only to chosen responses. Rejected responses may also contain shortcut cues that affect reward margins, and extending the framework to jointly probe both sides of a preference pair could provide a more complete characterization of shortcut reliance. Finally, we have not fully explored the relationship between shortcut reweighting, model capacity, and training dynamics. Stronger reward models may exhibit different sensitivity patterns, and shortcut sensitivity may change throughout optimization. Beyond these limitations, extending DynaCF to multimodal reward modeling~\cite{tian2026autorubricrewardimplicitpreferences,wang2025unifiedreward} and more expressive reward-model architectures~\cite{miao2026adajudge,wang2024interpretable} remains an important direction for future work.

\bibliography{custom}

\clearpage
\appendix


\section{Related Work}

\subsection{Reward Modeling for Preference Learning}

Reward modeling learns human preferences by assigning higher scalar scores to preferred responses, typically through pairwise Bradley--Terry objectives, and serves as a core component of RLHF, rejection sampling, reranking, and preference optimization pipelines~\cite{bradley1952rank,christiano2017deep,ziegler2019fine,stiennon2020learning,ouyang2022training,bai2022training,schulman2017proximal,zheng2023secrets,dong2024rlhf,rafailov2023direct}. Recent work has advanced reward modeling through stronger model backbones, larger preference datasets, improved evaluation benchmarks, process-level supervision, generative reward models, and more robust optimization objectives~\cite{yang2025qwen3,wang2025helpsteer3,liu2024rmbench,lambert2024rewardbench,malik2025rewardbench2,lightman2023lets,wang2024mathshepherd,mahan2024generative,wang2024secrets}. These developments substantially improve preference modeling capabilities, but also increase the importance of robustness and reliability, since reward models are often directly used to guide downstream policy optimization.

\subsection{Shortcut Learning in Reward Models}

Shortcut learning occurs when models rely on spurious correlations that are predictive in the training distribution but misaligned with the intended task~\cite{geirhos2020shortcut}. In reward modeling, shortcut behavior often emerges from superficial attributes such as response length, formatting patterns, verbosity, confidence markers, or template-like expressions, causing models to favor stylistic cues rather than genuine response quality~\cite{singhal2023long,shen2023loose,park2024disentangling,dubois2024length,wen2024language}. Because reward models are frequently used to optimize downstream policies, such shortcut biases can be amplified during training and lead to reward hacking or overoptimization~\cite{skalse2022defining,gao2023scaling,eisenstein2023helping,coste2024reward,rame2024warm,moskovitz2024confronting}. Prior work has explored robustness-oriented solutions including ensembles, weight averaging, disentangled rewards, causal reward modeling, and explicit robust objectives~\cite{eisenstein2023helping,coste2024reward,rame2024warm,chen2024odin,srivastava2025crome,wang2025beyond,liu2025rrm}. However, many existing approaches rely on static heuristics or predefined shortcut features, highlighting the need for more adaptive and model-aware shortcut diagnosis during training.

\section{Training Details}
\label{sec:appendix_training_details}

DynaCF uses a short dynamic warmup stage before applying counterfactual reweighting. Specifically, the first 5\% of training steps use the standard Bradley--Terry loss without shortcut-based downweighting. After warmup, each valid counterfactual sensitivity score is converted into a soft sample weight with reweighting strength $\gamma=1.0$ and minimum weight $w_{\min}=0.1$, so shortcut-sensitive samples are downweighted but never removed completely.

The overall training and LoRA hyperparameters are summarized in Table~\ref{tab:appendix_training_details}. Both reward models are trained for one epoch with global batch size 32, learning rate $4.0{\times}10^{-5}$, bf16 precision, and gradient checkpointing. We use LoRA fine-tuning~\cite{hu2021lora} for both backbones, with different LoRA ranks and context lengths for Qwen3-4B and Qwen3-8B. All training runs are conducted on $8\times$ NVIDIA RTX 5090 GPUs.

\begin{center}
\centering
\footnotesize
\setlength{\tabcolsep}{4pt}
\begin{tabular}{lcc}
\toprule
\textbf{Setting} & \textbf{Qwen3-4B} & \textbf{Qwen3-8B} \\
\midrule
Max length & 2048 & 4096 \\
Epochs & 1 & 1 \\
Global batch size & 32 & 32 \\
Learning rate & $4.0{\times}10^{-5}$ & $4.0{\times}10^{-5}$ \\
LR warmup ratio & 0.03 & 0.03 \\
Weight decay & 0.0 & 0.0 \\
Max grad norm & 1.0 & 1.0 \\
Precision & bf16 & bf16 \\
Gradient checkpointing & True & True \\
LoRA rank $r$ & 16 & 96 \\
LoRA alpha & 32 & 128 \\
LoRA dropout & 0.05 & 0.05 \\
DynaCF warmup ratio & 0.05 & 0.05 \\
Reweighting $\gamma$ & 1.0 & 1.0 \\
Minimum weight $w_{\min}$ & 0.1 & 0.1 \\
\bottomrule
\end{tabular}
\captionof{table}{Training and LoRA hyperparameters for the Qwen3 reward models.}
\label{tab:appendix_training_details}
\end{center}

\section{Additional Experiments}
\label{sec:appendix_additional_experiments}

This section provides additional ablations on DynaCF design choices. Unless otherwise noted, all experiments use Qwen3-4B with the same training setup as the main results and report benchmark-level overall scores. These studies examine the effects of reweighting hyperparameters, warmup scheduling, the diagnostic-only use of counterfactuals, and domain-aware counterfactual construction.

\begin{figure}[t]
\centering
\includegraphics[width=\columnwidth]{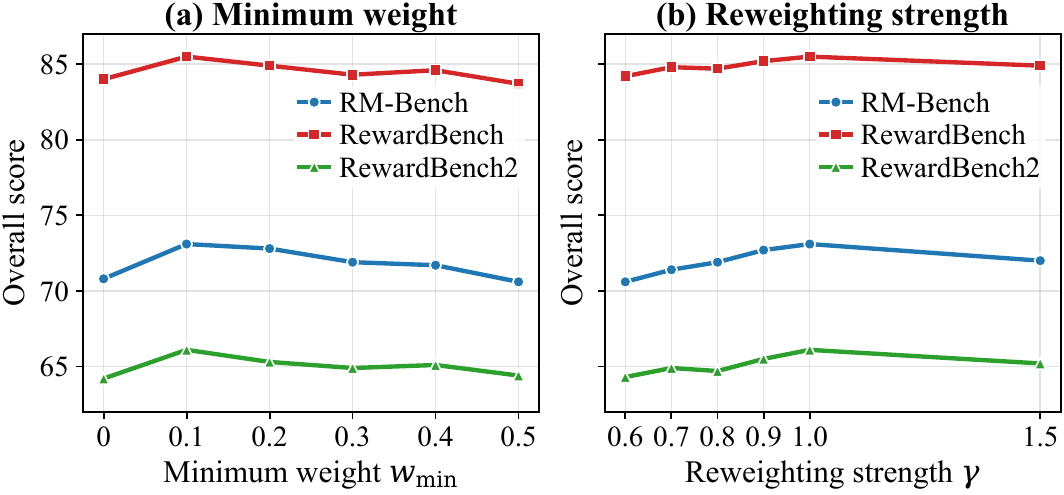}
\caption{Ablation results on Qwen3-4B. Left: benchmark-level overall scores under different minimum weights. Right: benchmark-level overall scores under different reweighting strengths.}
\label{fig:ablation_min_weight_gamma}
\end{figure}

\paragraph{Minimum weight.}
Figure~\ref{fig:ablation_min_weight_gamma} shows that performance first improves and then declines as $w_{\min}$ increases, supporting a small positive lower bound that attenuates shortcut-sensitive examples without discarding them. This trend suggests that retaining limited supervision from these examples is still beneficial for generalization.

\paragraph{Reweighting strength.}
Figure~\ref{fig:ablation_min_weight_gamma} also shows that moderate $\gamma$ yields the best trade-off, confirming that reweighting should suppress shortcut-sensitive updates while preserving useful preference supervision. Both weaker and stronger settings lead to less favorable overall behavior, consistent with the need for balanced regularization.

\paragraph{Dynamic warmup.}
We vary the fraction of early training steps that use the vanilla Bradley--Terry loss before enabling counterfactual reweighting. As shown in Table~\ref{tab:appendix_warmup_ablation}, a short warmup yields the best performance: enabling reweighting immediately can rely on noisy early sensitivity estimates, whereas a longer warmup delays shortcut mitigation.

\begin{center}
\centering
\footnotesize
\setlength{\tabcolsep}{3.5pt}
\begin{tabular}{@{}lcccc@{}}
\toprule
\textbf{Warmup} & \textbf{\shortstack{RM-\\Bench}} & \textbf{\shortstack{Reward\\Bench}} & \textbf{\shortstack{Reward\\Bench2}} & \textbf{Avg.} \\
\midrule
$0$ & 72.0 & 84.9 & 65.2 & 74.0 \\
$0.025$ & 72.7 & 85.2 & 65.7 & 74.5 \\
$0.05$ & \textbf{73.1} & \textbf{85.5} & \textbf{66.1} & \textbf{74.9} \\
$0.10$ & 72.4 & 85.0 & 65.4 & 74.3 \\
\bottomrule
\end{tabular}
\captionof{table}{Dynamic warmup ablation on Qwen3-4B.}
\label{tab:appendix_warmup_ablation}
\end{center}

\paragraph{Auxiliary counterfactual loss.}
We compare the diagnostic-only design with variants that additionally optimize a Bradley--Terry loss on counterfactual pairs. Table~\ref{tab:appendix_aux_cf_loss} shows that directly training on counterfactual pairs consistently underperforms the diagnostic-only design. This supports our choice to use counterfactuals as probes for measuring shortcut sensitivity, rather than as additional preference labels.

\begin{center}
\centering
\footnotesize
\setlength{\tabcolsep}{0pt}
\begin{tabular*}{\columnwidth}{@{\extracolsep{\fill}}lcccc@{}}
\toprule
\textbf{Variant} & \textbf{\shortstack{RM-\\Bench}} & \textbf{\shortstack{Reward\\Bench}} & \textbf{\shortstack{Reward\\Bench2}} & \textbf{Avg.} \\
\midrule
Diagnostic only & \textbf{73.1} & \textbf{85.5} & \textbf{66.1} & \textbf{74.9} \\
Aux CF loss, $\lambda=0.1$ & 72.6 & 85.0 & 65.7 & 74.4 \\
Aux CF loss, $\lambda=0.3$ & 71.8 & 84.5 & 65.1 & 73.8 \\
\bottomrule
\end{tabular*}
\captionof{table}{Auxiliary counterfactual loss ablation on Qwen3-4B.}
\label{tab:appendix_aux_cf_loss}
\end{center}

\paragraph{Counterfactual rule ablation.}
We ablate the rule profiles used to construct counterfactual samples. Table~\ref{tab:appendix_rule_ablation} shows that generic default rules already provide useful shortcut probes, while adding math- and code-specific rules improves the corresponding domains. The full domain-aware configuration gives the strongest overall RM-Bench result, indicating that counterfactuals should respect domain-specific structure rather than applying a single generic transformation set to all samples.

\begin{center}
\centering
\footnotesize
\setlength{\tabcolsep}{3pt}
\begin{tabular}{@{}lccccc@{}}
\toprule
\textbf{Rules} & \textbf{Chat} & \textbf{Math} & \textbf{Code} & \textbf{Safety} & \textbf{Overall} \\
\midrule
Default only       & 70.4 & 67.9 & 62.8 & 84.9 & 71.5 \\
Default + math     & 70.7 & 69.1 & 62.9 & 86.5 & 72.3 \\
Default + code     & 70.8 & 68.0 & 63.5 & 85.7 & 72.0 \\
Full domain-aware  & \textbf{71.2} & \textbf{69.2} & \textbf{63.4} & \textbf{88.7} & \textbf{73.1} \\
\bottomrule
\end{tabular}
\captionof{table}{Counterfactual rule ablation on RM-Bench with Qwen3-4B. Overall is the average of Chat, Math, Code, and Safety.}
\label{tab:appendix_rule_ablation}
\end{center}

\section{Computational Cost Analysis}
\label{sec:appendix_compute_cost}

We report the computational efficiency of DynaCF using throughput measured in TFLOPs per hour. As shown in Table~\ref{tab:appendix_tflops}, this analysis quantifies the additional cost introduced by counterfactual sensitivity computation and dynamic reweighting relative to standard Bradley--Terry training. The main overhead comes from scoring valid counterfactual responses; the reweighting operation itself is lightweight because it only rescales the original Bradley--Terry loss.

\begin{center}
\centering
\footnotesize
\setlength{\tabcolsep}{3pt}
\begin{tabular}{@{}lccc@{}}
\toprule
\textbf{Model} & \textbf{Training} & \textbf{TFLOPs/h} & \textbf{Relative Cost} \\
\midrule
Qwen3-4B & Bradley--Terry & 132.4 & 1.00$\times$ \\
Qwen3-4B & DynaCF & 118.7 & 1.12$\times$ \\
Qwen3-8B & Bradley--Terry & 109.6 & 1.00$\times$ \\
Qwen3-8B & DynaCF & 96.8 & 1.13$\times$ \\
\bottomrule
\end{tabular}
\captionof{table}{Computational cost measured by TFLOPs per hour.}
\label{tab:appendix_tflops}
\end{center}

\section{DynaCF Training Procedure}
\label{sec:appendix_dynacf_algorithm}

Algorithm~\ref{alg:dynacf} gives the complete training procedure for DynaCF. At each training step, the current reward model scores both the original preference pair and the valid counterfactual variants of the chosen response. The resulting margin changes and preference flips are converted into a counterfactual shortcut-sensitivity score, which is then mapped to a step-local sample weight for the original Bradley--Terry objective. Thus, counterfactuals serve as diagnostic probes rather than additional preference labels, and the original preference distribution remains unchanged during optimization.
\begin{table*}[t]
\centering
\footnotesize
\setlength{\tabcolsep}{6pt}
\begin{tabular}{@{}>{\raggedright\arraybackslash}p{0.18\textwidth}
                >{\raggedright\arraybackslash}p{0.30\textwidth}
                >{\raggedright\arraybackslash}p{0.45\textwidth}@{}}
\toprule
\textbf{Type} & \textbf{Surface cue} & \textbf{Counterfactual operation} \\
\midrule
Length 
& Longer chosen response 
& Shorten redundant text while preserving the main answer \\

Format 
& Markdown, bullets, headings 
& Normalize headings, lists, emphasis, and spacing \\

Verbosity 
& Repeated explanations or fillers 
& Remove repetition, summaries, and non-essential phrases \\

Tone 
& Confident or template-like wording 
& Neutralize overly polished or formulaic expressions \\
\bottomrule
\end{tabular}
\caption{Shortcut categories and counterfactual operations used in the analysis.}
\label{tab:appendix_shortcut_taxonomy}
\end{table*}

\begin{table*}[t]
\centering
\footnotesize
\setlength{\tabcolsep}{6pt}
\begin{tabular}{@{}p{0.18\textwidth}p{0.30\textwidth}p{0.33\textwidth}@{}}
\toprule
\textbf{Check} & \textbf{Rule} & \textbf{Purpose} \\
\midrule
Non-degenerate edit & Non-empty and not identical & Remove failed or no-op transformations \\
Default overlap & At least $0.55$ token overlap & Preserve main semantic content \\
Math overlap & At least $0.85$ token overlap & Protect equations and derivations \\
Code overlap & At least $0.75$ token overlap & Preserve code logic and interfaces \\
Number preservation & Retain all numbers for math/code & Avoid corrupting structured content \\
Length gate & Shorten only when chosen is longer & Target plausible length shortcuts \\
CF cap & Max CFs: default 4, math 4, code 6 & Bound scoring overhead per sample \\
Fallback & No valid CF gives $s_{\mathrm{cf}}=0$ & Avoid penalizing uncertain probes \\
\bottomrule
\end{tabular}
\caption{Validity checks for counterfactual samples.}
\label{tab:appendix_cf_validity}
\end{table*}

\begin{algorithm}[H]
\caption{Dynamic Counterfactual Reweighting}
\label{alg:dynacf}
\small
\begin{algorithmic}[1]
\Require Batch $B_t$, current RM $R_{\theta_t}$, counterfactual cache $\mathcal{C}$
\Ensure Weighted BT loss $\mathcal{L}_{\mathrm{DynaCF}}(t)$

\For{each sample $i \in B_t$}
    \State Compute original margin $m_i(t)$
    \State Retrieve or construct valid counterfactuals $\{y_{i,k}^{\mathrm{cf}}\}_{k=1}^{K_i}$
    \State Compute counterfactual margins $m_{i,k}^{\mathrm{cf}}(t)$
    \State Compute shortcut sensitivity $s_{\mathrm{cf},i}(t)$ using Eq.~\ref{eq:counterfactual-sensitivity}
    \State Convert $s_{\mathrm{cf},i}(t)$ into weight $w_i(t)$ using Eq.~\ref{eq:dynamic-weight}
\EndFor

\State Compute $\mathcal{L}_{\mathrm{DynaCF}}(t)$ using Eq.~\ref{eq:dynamic-objective}
\State Update parameters with learning rate $\eta$:
\Statex \hspace{\algorithmicindent}$\theta_{t+1}\leftarrow\theta_t-\eta\nabla_{\theta_t}\mathcal{L}_{\mathrm{DynaCF}}(t)$
\end{algorithmic}
\end{algorithm}

\section{Shortcut Types}
\label{sec:appendix_shortcut_types}

Table~\ref{tab:appendix_shortcut_taxonomy} summarizes the shortcut categories used in the analysis and the corresponding counterfactual operations used to probe each category.


\section{Counterfactual Sample Design}
\label{sec:appendix_counterfactual_design}

Counterfactual construction changes surface cues while preserving task-relevant content. We use three rule profiles: \textit{default}, \textit{math}, and \textit{code}. Chat and safety examples use the default profile, while math and code examples use domain-specific profiles to avoid corrupting structured content.

\paragraph{Normal profile.}
The normal profile targets general presentation shortcuts in dialogue and safety examples. It normalizes formatting, reduces unnecessary length, neutralizes overly polished tone, and removes redundant verbosity. These transformations are intended to alter how the answer is presented without changing its main semantic content.

\begin{itemize}[leftmargin=*, itemsep=0pt, parsep=0pt, topsep=0pt]
    \item \textbf{Formatting}: remove Markdown emphasis, headings, bullets, numbering, and excessive blank lines.
    \item \textbf{Length}: shorten redundant chosen responses when the chosen answer is substantially longer than the rejected answer.
    \item \textbf{Tone}: replace confident, flattering, or template-like phrases with neutral wording.
    \item \textbf{Verbosity}: remove repeated sentences, redundant summaries, and filler phrases.
\end{itemize}

\paragraph{Math profile.}
The math profile is conservative because numbers, equations, and derivations are often task-relevant. It removes non-essential greetings, headings, emphasis, and stylistic preambles while preserving equations, numerical values, reasoning steps, and final answers. Valid math counterfactuals must satisfy a high token-overlap threshold and preserve all numbers.

\begin{itemize}[leftmargin=*, itemsep=0pt, parsep=0pt, topsep=0pt]
    \item \textbf{Tone neutralization}: neutralize wording without changing mathematical content.
    \item \textbf{Light formatting normalization}: remove headings or emphasis while preserving equations and derivations.
    \item \textbf{Preamble removal}: remove greetings, setup phrases, and non-essential lead-in text before the solution.
    \item \textbf{Style normalization}: combine tone neutralization, preamble removal, and light formatting normalization.
\end{itemize}

\paragraph{Code profile.}
The code profile targets presentation cues around code while preserving executable logic. It keeps code blocks, interfaces, variables, control flow, and numbers intact, and only modifies surrounding explanations or non-essential surface markers.

\begin{itemize}[leftmargin=*, itemsep=0pt, parsep=0pt, topsep=0pt]
    \item \textbf{Explanation reduction}: reduce prose outside fenced code blocks while keeping code intact.
    \item \textbf{Comment neutralization}: remove obvious or verbose comments while preserving warnings, invariants, and edge-case notes.
    \item \textbf{Wrapper removal}: remove phrases such as ``here is the solution'' and Markdown list markers around code.
    \item \textbf{Readability normalization}: normalize trailing spaces and excessive blank lines.
    \item \textbf{Claim removal}: remove unsupported efficiency or complexity claims outside the code.
    \item \textbf{Test presentation}: reduce test/example presentation when it acts as a surface cue rather than core logic.
\end{itemize}

\begin{algorithm}[H]
\caption{Counterfactual Validity Checks}
\label{alg:appendix_validity_checks}
\footnotesize
\begin{algorithmic}[1]
\Require Chosen response $y_i^+$, candidates $\mathcal{C}_i$, domain $d$
\Ensure Valid counterfactual set $\mathcal{V}_i$
\State Set $\mathcal{V}_i \leftarrow \emptyset$ and load $\tau_d,p_d$
\For{each $y_{i,k}^{\mathrm{cf}}\in\mathcal{C}_i$}
    \State \textbf{skip if} $y_{i,k}^{\mathrm{cf}}$ is empty or identical to $y_i^+$
    \State \textbf{skip if} $\mathrm{TokenOverlap}(y_i^+,y_{i,k}^{\mathrm{cf}})<\tau_d$
    \State \textbf{skip if} $p_d$ and numbers in $y_i^+$ are not preserved
    \State Add $y_{i,k}^{\mathrm{cf}}$ to $\mathcal{V}_i$
\EndFor
\If{$\mathcal{V}_i=\emptyset$}
    \State Set $s_{\mathrm{cf},i}=0$ and keep full sample weight
\EndIf
\end{algorithmic}
\end{algorithm}

\paragraph{Validity checks.}
All generated counterfactuals are filtered before shortcut sensitivity computation. The checks are designed to reject degenerate edits, preserve the task-relevant content of the chosen response, and avoid assigning shortcut penalties when no reliable probe is available. Token overlap serves as the main semantic-preservation criterion, with stricter thresholds for math and code because small edits can change correctness. For structured domains, all numbers in the original response must be preserved, which helps protect equations, constants, test cases, and boundary conditions. Length-reduction counterfactuals are only generated when the chosen response is substantially longer than the rejected response, so the operation targets plausible length shortcuts rather than arbitrary compression. The filtering procedure is shown in Algorithm~\ref{alg:appendix_validity_checks}, and the concrete checks are summarized in Table~\ref{tab:appendix_cf_validity}. If no valid counterfactual remains, the sample receives zero shortcut sensitivity and keeps its original training weight.


\end{document}